\documentclass[letterpaper]{article} 
\usepackage[preprint]{aaai2027}  
\usepackage{times}  
\usepackage{helvet} 
\usepackage{courier}  
\usepackage[hyphens]{url}  
\usepackage{graphicx} 
\usepackage{natbib}  
\usepackage{caption} 
\usepackage{amsmath}
\usepackage{amssymb}
\usepackage{booktabs}

\usepackage{siunitx}

\title{R-GEAN: Regimen-Guided Edit Action Network for Within-Admission Medication Change Prediction}

\author{
    Regan Mahat\textsuperscript{\rm 1},
    Mansu Kim\textsuperscript{\rm 1}
}

\affiliations{
    \textsuperscript{\rm 1}Gwangju Institute of Science and Technology (GIST),
    Gwangju, Republic of Korea\\
    Regan Mahat: reganmahat@gm.gist.ac.kr\\
    Mansu Kim: mansu.kim@gist.ac.kr
}

\begin{document}

\maketitle

\begin{abstract}
The medications prescribed to a patient often change during a hospital
admission as clinicians start, stop, or continue therapies. We study whether models can predict which medication classes are added or
removed between 24 hours after admission and discharge. Metrics that compare the complete discharge regimen can reward models for copying medications that remain unchanged, even when they identify no actual changes. We therefore introduce a leakage-controlled benchmark that predicts net ATC3 additions and removals using only prior completed admissions and information available within the first 24 hours of the current admission. Addition candidates are classes not active at 24 hours, whereas removal candidates are classes active at that time.

We also introduce R-GEAN, an asymmetric candidate-scoring network with independent addition and removal predictors. Across 240,480 admissions from 82,286 patients, R-GEAN achieves the highest predefined summary of addition, removal, changed-regimen, and action-pattern performance, termed the edit composite (0.464), compared with 0.435 for the strongest primary comparator. Reimplemented RETAIN, GAMENet, and MICRON baselines obtain 0.428, 0.420, and 0.288, respectively. R-GEAN's advantage is concentrated in correctly identifying medication classes no longer active at discharge, while rare additions and admissions with multiple medication changes remain difficult. Rankings based on micro-F1 over the reconstructed discharge regimen and the edit composite correlate weakly across the evaluated models (Spearman $\rho=0.20$). The continuation baseline achieves the highest complete-regimen score despite predicting no additions or removals. These results show that complete-regimen and edit-level evaluation measure different aspects of medication prediction. The benchmark evaluates observed prescribing changes, not treatment appropriateness.
\end{abstract}

\section{Introduction}

Medication treatment often changes during a hospital admission. Clinicians may start, stop, or continue therapies as the patient's condition and treatment needs evolve \citep{daliri2021longitudinal}. As a result, the medication classes active early in an admission may differ from those active at discharge. Our goal is to predict these changes: which medication classes will be added and which will be removed by discharge.

For each admission, we observe the medication classes active after 24 hours and call this set the \emph{anchor regimen}. The task predicts two net changes relative to this observed regimen: additions are classes absent at 24 hours but active at discharge, whereas removals are classes active at 24 hours but inactive at discharge. These labels compare the 24-hour and discharge snapshots and do not represent every medication event occurring between them. To prevent temporal leakage, models use only prior completed admissions and information available within the first 24 hours of the current admission \citep{pungitore2023timewindows}.

Complete-set evaluation can hide a model's failure to predict medication changes. A complete-set metric compares the entire predicted discharge regimen with the observed regimen, whereas an edit-level metric evaluates additions and removals directly. Most medication-prediction systems predict the complete medication set and evaluate it using Jaccard similarity or micro-averaged F1 \citep{zhang2017leap,choi2016retain,shang2019gamenet,yang2021safedrug}. Because many therapies remain unchanged, a model can obtain a high complete-set score by copying the 24-hour regimen without identifying any actual changes \citep{yang2021micron}. In our benchmark, this continuation baseline attains the highest micro-F1 over the reconstructed discharge regimen, 0.8169, but only 0.1634 on the edit composite because it predicts no additions or removals. We therefore report both complete-set metrics, which credit correctly continued therapies, and edit-level metrics, which directly measure addition and removal recovery. We also test whether these two evaluation views rank models differently.

R-GEAN predicts additions and removals with separate predictors because the two decisions rely on different evidence. The addition predictor scores medication classes not present after 24 hours using the current clinical state, whereas the removal predictor scores classes already present using medication-specific exposure history. Their predictions are combined to reconstruct the medication regimen active at discharge.

Our contributions are as follows.

\begin{itemize}
\item A leakage-controlled benchmark that separates medication additions and removals from therapies that remain unchanged. It enforces a 24-hour information boundary and provides patient-disjoint splits and change-recovery metrics for 240,480 MIMIC-IV admissions from 82,286 patients and 78 ATC3 classes. The task, cohort, and leakage controls were fixed before model training and verified by automated tests.
\item R-GEAN, an asymmetric candidate-scoring model with separate predictors for additions and removals. The model scores only classes eligible for each direction and uses current clinical information and medication-specific exposure history. R-GEAN achieves the highest edit composite, with its advantage concentrated in removal prediction and supported by Holm-corrected patient-level bootstrap analysis.
\item An evaluation showing that complete-set and edit-level metrics rank models differently, motivating the use of both evaluation views.
\end{itemize}

\section{Related Work}

\paragraph{Complete-set medication recommendation.}
Most prior work predicts the full medication regimen for each visit rather than
changes from an observed regimen. Existing methods use current-visit
information, longitudinal histories, molecular structure, generative decoding,
substructure-aware representations, or selective retrieval of prior visits
\citep{zhang2017leap,choi2016retain,shang2019gamenet,yang2021safedrug,
wu2022cognet,yang2023molerec,kim2024vita}. We report complete-set metrics for
comparison with this literature and edit-level metrics for direct change
recovery.

\paragraph{Medication change prediction.}
MICRON is the closest prior method because it also predicts medication
additions and removals \citep{yang2021micron}. It models visit-to-visit changes
from a maintained medication vector, whereas R-GEAN predicts within-admission
changes from an observed 24-hour regimen. R-GEAN also restricts additions to
classes absent at 24 hours and removals to classes present at that time. MICRON
was developed on MIMIC-III, while R-GEAN is evaluated on MIMIC-IV
\citep{johnson2016mimiciii,johnson2023mimiciv}. We reimplement MICRON on this
benchmark; implementation details and the DualNN comparison are provided in the
supplement.

\paragraph{Evaluation of medication recommendation.}
To our knowledge, prior work has reported addition and removal errors
\citep{yang2021micron}, but has not directly tested whether models receive the
same ranking under complete-set and edit-level evaluation.

\section{Task and Benchmark}
\label{sec:task}

\subsection{Formulation}

The task compares the medication classes active 24 hours after admission
with those active at discharge. Let $\mathcal{V}$ denote the vocabulary of
$|\mathcal{V}|=78$ ATC3 therapeutic classes. For each admission, let $M_A \subseteq \mathcal{V}$ be the medication classes
recorded as active 24 hours after admission; we call this the
\emph{anchor regimen}. Let $M_T \subseteq \mathcal{V}$ be the medication
classes recorded as active at discharge; we call this the
\emph{target regimen}. Activity at discharge is determined from recorded
prescription start and stop times. Because MIMIC-IV does not provide a
reconciled discharge medication list, $M_T$ is an approximation of the
medication classes active at discharge.

The prediction targets are the two net changes between these medication sets:
\begin{equation}
Y^{+} = M_T \setminus M_A,
\qquad
Y^{-} = M_A \setminus M_T ,
\label{eq:targets}
\end{equation}
Here, $Y^{+}$ contains additions: classes absent at 24 hours but active at
discharge. Similarly, $Y^{-}$ contains removals: classes active at 24 hours but
inactive at discharge. These targets compare the two medication snapshots and
do not represent every event occurring between them. In particular, they
exclude temporary exposures that begin and end between the landmarks and
stop--restart sequences that do not change snapshot membership.

A model predicts $\hat{Y}^{+}$ and $\hat{Y}^{-}$ and reconstructs the target
regimen as
\begin{equation}
\hat{M}_T = \left( M_A \setminus \hat{Y}^{-} \right) \cup \hat{Y}^{+} .
\label{eq:reconstruction}
\end{equation}

For each prediction direction, the candidate space is the set of medication
classes that can validly be predicted. Addition candidates are classes absent
from the anchor regimen, whereas removal candidates are classes present in it:
\begin{equation}
\mathcal{C}^{+} = \mathcal{V} \setminus M_A ,
\qquad
\mathcal{C}^{-} = M_A .
\label{eq:candidates}
\end{equation}
These candidate spaces cover all benchmark labels in the final 78-class
vocabulary and are disjoint, so
$\hat{Y}^{+} \cap \hat{Y}^{-} = \emptyset$ by construction.

\begin{figure}[t]
\centering
\includegraphics[width=0.98\columnwidth]{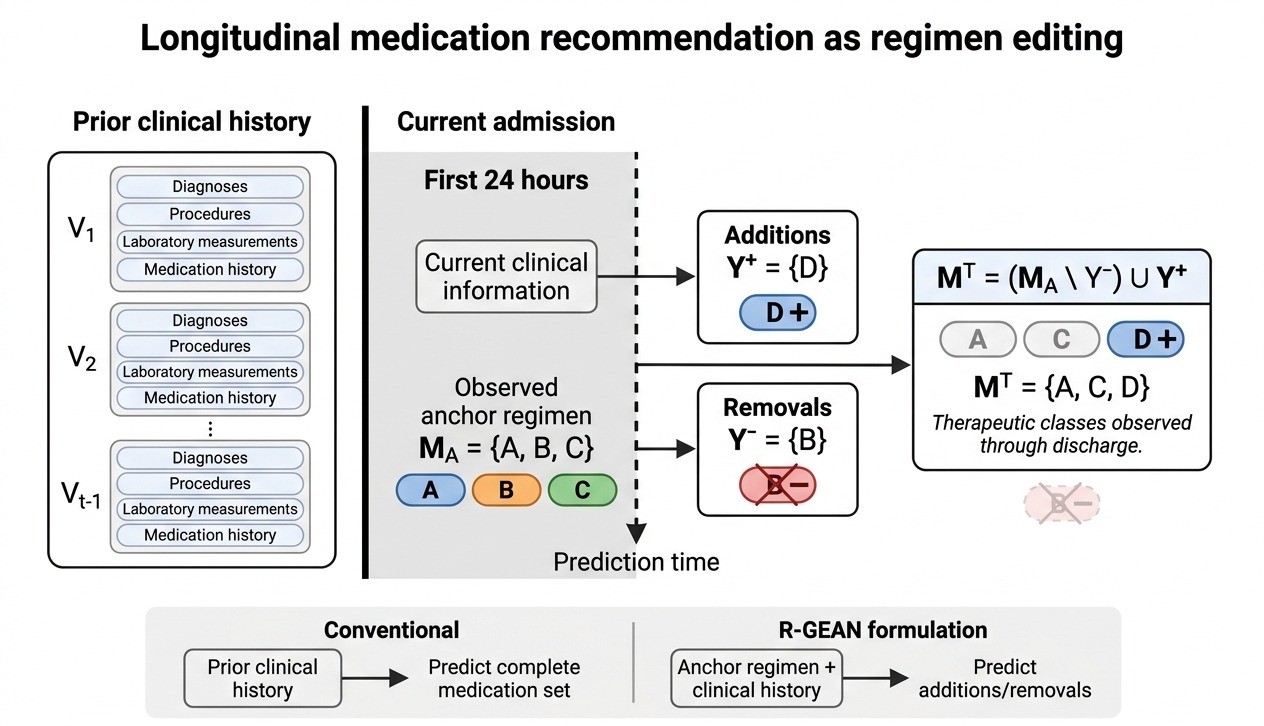}
\caption{The model uses prior history and information available within the first
24 hours to predict medication classes added to or removed from the observed
24-hour regimen by discharge. The two predictions are combined to reconstruct
the discharge regimen. No information recorded after the 24-hour boundary is
available to the model.}
\label{fig:task}
\end{figure}

\subsection{Cohort and construction}

The benchmark includes adults from MIMIC-IV v3.1 with at least one completed
admission before the current hospital stay
\citep{johnson2023mimiciv,physionet2024mimiciv31}. An admission is eligible if
the patient is at least 18 years old, the admission and discharge times are
valid, and prescription and diagnosis records are present. Earlier admissions
provide patient history, whereas the 24-hour medication regimen is constructed
only from the current admission.

Cohort and admission data come from the MIMIC-IV v3.1 HOSP module. Laboratory
measurements come from \texttt{hosp/labevents}, and vital-sign change features
are derived from linked \texttt{icu/chartevents} records when available.
Admissions without linked vital-sign records remain in the cohort and receive
the corresponding no-data states.

This design excludes first admissions and patients with only one admission,
which limits generalizability to those populations. Compared with otherwise
eligible adults with a single admission, the retained patients are older
(62.9 vs.\ 56.6 years; standardized mean difference 0.33) and have longer
hospital stays.

Patients are divided into training, validation, and test sets in a
$70{:}10{:}20$ ratio using a seeded random permutation of patient identifiers
(seed 2026). All admissions from the same patient remain in one split, ensuring
that the partitions are patient-disjoint.

We convert medication records into a fixed vocabulary of 78 ATC3 therapeutic
classes. Medication names are first standardized with RxNorm
\citep{nelson2011rxnorm,zeng2006rxnav} and then mapped to the Anatomical
Therapeutic Chemical system \citep{who_atc_classification}. We use ATC level 3
to reduce sparsity while retaining clinically meaningful therapeutic groups.
After excluding out-of-scope records under the prespecified policy, 99.4\% of
the remaining eligible medication records are resolved. All 78 retained classes are represented in the training cohort. Detailed
mapping rules, intermediate coverage statistics, and handling of medications
with multiple possible ATC codes are provided in the supplement.

We defined the 24-hour cutoff, cohort eligibility criteria, discharge target, and medication-mapping rules before training any model and kept them unchanged throughout model development. Table~\ref{tab:benchmark} summarizes the final benchmark.

\begin{table}[t]
\centering
\setlength{\tabcolsep}{4pt}
\begin{tabular}{lr}
\toprule
\textbf{Property} & \textbf{Value} \\
\midrule
Source & MIMIC-IV v3.1 \\
Patients & 82,286 \\
Admission-level examples & 240,480 \\
Target vocabulary (ATC3) & 78 classes \\
\midrule
Train examples / patients & 167,418 / 57,627 \\
Validation examples / patients & 24,030 / 8,215 \\
Test examples / patients & 49,032 / 16,444 \\
\bottomrule
\end{tabular}
\caption{Benchmark and split statistics. The splits are patient-disjoint. Each
test admission belongs to one of seven mutually exclusive medication-change
patterns: \textsc{continue} 9,580, \textsc{add} 9,879, \textsc{remove} 7,818,
\textsc{switch} 7,389, \textsc{multi-edit} 13,560,
\textsc{empty-to-nonempty} 749, and \textsc{nonempty-to-empty} 57.}
\label{tab:benchmark}
\end{table}

\subsection{Leakage control}

We control leakage through patient-disjoint splits, a fixed 24-hour information
boundary, training-only feature construction, validation-only threshold
selection, a locked test evaluation, and direction-specific candidate masking.
All admissions from a patient remain in one split, and model inputs are limited
to prior completed admissions and information available within the first
24 hours of the current admission. Data-derived statistics, including
co-prescription information, are computed from the training split only.
Decision thresholds are selected on the full validation candidate set. Primary
models, thresholds, and confirmatory comparisons are fixed before test
evaluation; later ablations and diagnostics are reported as secondary analyses
on the same test split. Candidate masking follows
Equation~\ref{eq:candidates}, so models score only classes valid for the
corresponding prediction direction.

Clinical measurements are included only when their recorded availability time
falls within the first 24 hours. Laboratory and vital-sign values require
\texttt{storetime} at or before the cutoff, excluding measurements collected
earlier but recorded later. Medication-exposure features use only orders
available before the cutoff, and a medication is marked as discontinued only
when its recorded stop time is at or before 24 hours. We verified across the
full cohort that every R-GEAN feature block satisfies these availability rules,
and all automated leakage tests passed. Detailed feature provenance is provided
in the supplement.

\section{R-GEAN}
\label{sec:model}

\subsection{Design}

\begin{figure*}[t]
\centering
\includegraphics[width=0.98\textwidth]{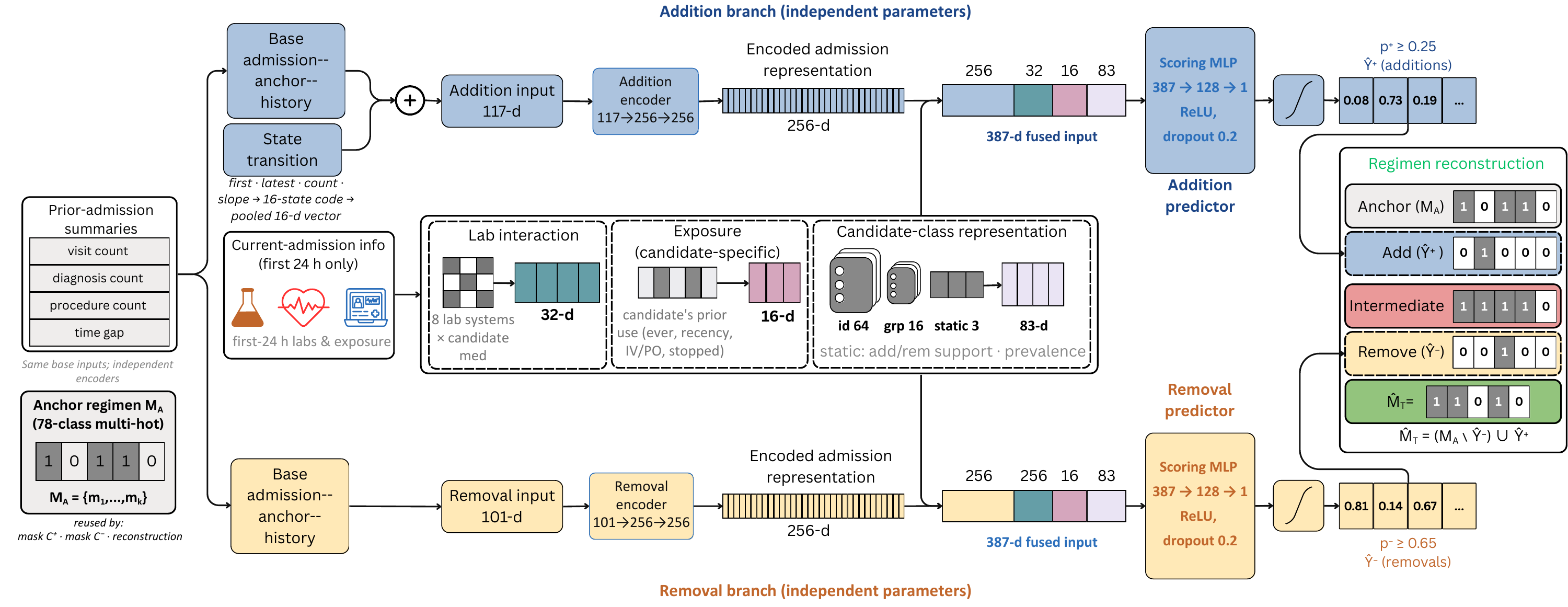}
\caption{R-GEAN architecture. The addition and removal encoders receive 117- and 101-dimensional admission inputs, respectively, and produce 256-dimensional admission representations. Each representation is combined with a 32-dimensional medication-specific laboratory vector, a 16-dimensional exposure vector, and an 83-dimensional medication-class representation. The resulting 387-dimensional vector is scored by a $387\!\rightarrow\!128\!\rightarrow\!1$ MLP. The addition branch evaluates classes absent from the observed 24-hour regimen ($\mathcal{C}^{+}=\mathcal{V}\setminus M_A$), whereas the removal branch evaluates classes present in it ($\mathcal{C}^{-}=M_A$). Their predictions reconstruct the discharge regimen as $\hat{M}_T=(M_A\setminus\hat{Y}^{-})\cup\hat{Y}^{+}$.}
\label{fig:arch}
\end{figure*}

R-GEAN predicts additions and removals separately because the two decisions use
different evidence. Predicting an addition depends mainly on the patient's
current clinical state and changes observed during the first 24 hours.
Predicting a removal depends more strongly on the patient's prior exposure to
that medication class. R-GEAN therefore uses separate neural predictors for
additions and removals. The two predictors are trained and decoded independently,
and their outputs are combined only when reconstructing the discharge regimen
(Figure~\ref{fig:arch}).

\subsection{Inputs and representations}

R-GEAN scores each candidate medication class by combining shared patient and
admission context with information specific to that class. The shared context
is a 101-dimensional vector containing 13 demographic, admission, and
prior-history features; a 78-dimensional multi-hot representation of the
medication regimen active at 24 hours, $M_A$; and 10 indicators of data
availability and healthcare use. The first group includes age, admission type,
and counts of prior visits, diagnoses, and procedures. Diagnoses and procedures
are represented as aggregate counts rather than code-level embeddings or visit
sequences.

The addition predictor also receives a patient-level summary of laboratory and
vital-sign changes during the first 24 hours. For each available variable, the
transition module assigns one of 16 ordered clinical-state codes using its first
and latest values, observation count, and slope relative to reference bands
computed from the training split. A shared embedding maps each code to a
16-dimensional vector, and the vectors are averaged across variables. This
overall transition representation extends the shared context from 101 to
117 dimensions for the addition encoder. The addition and removal encoders use
$117\!\rightarrow\!256\!\rightarrow\!256$ and
$101\!\rightarrow\!256\!\rightarrow\!256$ mappings, respectively.

Both predictors also use information specific to the medication class being
scored. This includes a 32-dimensional laboratory-interaction vector derived
from first-24-hour measurements and a 16-dimensional representation of
medication exposure available before the 24-hour cutoff. Each class is also
represented by a 64-dimensional identity embedding, a 16-dimensional
therapeutic-group embedding, and three class-level statistics computed from the
training split: addition support, removal support, and overall prevalence. The
identity embedding, therapeutic-group embedding, and three statistics together
form an 83-dimensional candidate representation.

For each candidate, the 256-dimensional admission representation is combined
with the 32-dimensional candidate-specific laboratory vector, 16-dimensional
exposure vector, and 83-dimensional candidate representation. The resulting
387-dimensional vector is scored by a
$387\!\rightarrow\!128\!\rightarrow\!1$ MLP.

The observed 24-hour regimen enters through its multi-hot representation, the
addition and removal candidate masks, and the discharge-regimen reconstruction
in Equation~\ref{eq:reconstruction}; no separate learned anchor encoder is used.

\subsection{Training and decoding}

The addition and removal tasks are highly imbalanced, with positive candidate
pairs substantially less frequent than negative pairs. To address this
imbalance, each predictor is trained using weighted binary cross-entropy. The
two predictors are optimized separately. The
positive-class weight is computed from the class distribution in the training
set:
\[
w_{+} =
\min\!\left(6,\max\!\left(1,\frac{n_{\mathrm{neg}}}{n_{\mathrm{pos}}}\right)\right),
\]
where the upper bound of 6 prevents extreme class weighting. Negative training pairs are subsampled to a 3:1 negative-to-positive ratio.

The addition predictor includes an auxiliary penalty for predicted medication pairs recorded as interacting in TWOSIDES. Let $A_{\mathrm{ddi}}$ denote the TWOSIDES interaction matrix mapped to the 78 ATC3 classes \citep{tatonetti2012twosides}. The regularization term $\Omega_{\mathrm{ddi}}$ represents the expected number of mapped interacting pairs formed between the predicted additions and the reconstructed discharge regimen. It is applied only to the addition predictor because adding a medication can introduce new interaction pairs, whereas removing a medication cannot introduce a new pair. The resulting training objectives are:
\begin{equation}
\mathcal{L}^{+}
=
\ell_{\text{wbce}}^{+}
+
0.01\,\Omega_{\mathrm{ddi}},
\qquad
\mathcal{L}^{-}
=
\ell_{\text{wbce}}^{-}.
\label{eq:loss}
\end{equation}

Both predictors are optimized with Adam \citep{kingma2015adam} using a learning
rate of $2\times10^{-3}$, weight decay of $10^{-5}$, a batch size of
65{,}536, and 16 training epochs.

At inference, a medication class is predicted as an addition or removal when
its score exceeds the corresponding direction-specific threshold:
\begin{equation}
\begin{split}
\hat{Y}^{+} &= \{m \in \mathcal{C}^{+}: p^{+}(m) \geq \tau^{+}\},\\
\hat{Y}^{-} &= \{m \in \mathcal{C}^{-}: p^{-}(m) \geq \tau^{-}\}.
\end{split}
\label{eq:decode}
\end{equation}
The addition and removal thresholds are selected independently by maximizing
the corresponding micro-F1 on the full validation candidate set over
$\{0.05,0.10,\ldots,0.95\}$. This procedure yields
$\tau^{+}=0.25$ and $\tau^{-}=0.65$. Validation candidates are not subsampled,
so threshold selection reflects the class distribution encountered at
inference. Predictions use the raw model probabilities without post-hoc
calibration; calibration is evaluated separately using ECE and Brier score
\citep{guo2017calibration,naeini2015calibration,brier1950verification}.

\section{Experimental Setup}

\subsection{Baselines}

We compare R-GEAN with baselines chosen to test three questions: whether
medication changes can be predicted without a learned model, whether existing
public models transfer to this task, and whether separate addition and removal predictors improve performance.

The non-learned baselines are \textsc{continuation}, which predicts no
medication changes, and \textsc{frequency}, which predicts changes using
class-level frequencies. The public baselines are RETAIN
\citep{choi2016retain}, GAMENet \citep{shang2019gamenet}, and MICRON
\citep{yang2021micron}. We use ``FullSet'' for models that predict the complete
discharge regimen and then derive additions and removals, and ``Edit'' for
variants that predict additions and removals directly. RETAIN and GAMENet are
evaluated in both forms. MICRON is reimplemented using its residual
reconstruction approach, interaction penalty, and separate addition and removal
thresholds.

The internal baselines are a gradient-boosted \textsc{tree} model and a
\textsc{shared} neural model. The tree baseline is a LightGBM classifier over
admission--candidate pairs \citep{ke2017lightgbm}. The shared model uses one
common encoder with separate addition and removal output heads and receives all
R-GEAN feature blocks together with a general clinical-summary input. It controls for feature access; a separately widened parameter-matched version
tests whether the advantage persists at comparable model capacity. LightGBM hyperparameters are provided in the supplement.

SafeDrug \citep{yang2021safedrug} relies on molecular information below the
ATC3 class level. A substantially adapted ATC3 version achieves a composite of
0.2816 and is reported only as a sensitivity analysis. It is excluded from the
primary comparison and rank-correlation analysis.

Comparisons with public models are interpreted at the system level because
their task adaptations and available inputs differ. The shared model provides
the controlled comparison for separate addition and removal predictors.

\subsection{Metrics}

Addition F1 ($\mathrm{F1}^{+}$) and removal F1
($\mathrm{F1}^{-}$) measure how accurately the corresponding medication
changes are recovered. Changed-admission Jaccard ($J_{\text{chg}}$) compares
the reconstructed and observed discharge regimens only for admissions with at
least one true addition or removal. Action macro-F1
($\mathrm{F1}_{\text{act}}$) measures whether the overall change pattern---such
as no change, addition, removal, switch, or multiple changes---is correctly
identified. The seven action groups are mutually exclusive and cover all test
admissions; their full definitions and precedence rules are provided in the
supplement.

We combine these four measures into a predefined edit composite used to
summarize model performance:
\begin{equation}
S = 0.35\,\mathrm{F1}^{+} + 0.30\,\mathrm{F1}^{-}
  + 0.25\,J_{\text{chg}} + 0.10\,\mathrm{F1}_{\text{act}}.
\label{eq:composite}
\end{equation}
The weights were fixed before final model selection: 65\% of the composite is
assigned to addition and removal F1, 25\% to changed-admission Jaccard, and
10\% to action macro-F1. The composite is a model-comparison summary rather than a clinical utility measure; we
therefore report all four components separately and do not interpret a higher
composite as evidence of clinical benefit.

We also report resulting-set micro-F1, which compares the reconstructed and
observed discharge regimens across all medication classes. This metric supports
comparison with methods that predict the complete medication set.

\subsection{Protocol}

The primary models, decision thresholds, and confirmatory comparisons were
specified before the locked test evaluation. Results in the main tables use
seed 2026. Later ablations and diagnostic analyses use the same test split and
are treated as secondary analyses.

R-GEAN remains the highest-ranked model across three independent training seeds
(2026--2028), with a mean edit composite of $0.4636 \pm 0.0047$ compared with
$0.4334 \pm 0.0041$ for the strongest competing model. Full seed-specific
results are provided in the supplement.

We use patient-level paired bootstrap sampling to estimate uncertainty while
preserving the dependence among admissions from the same patient. Holm
correction is applied jointly to eleven prespecified comparisons: four
edit-composite comparisons against RETAIN-FullSet, GAMENet-FullSet, the tree
model, and the shared model, and seven metric-specific comparisons against
MICRON \citep{holm1979simple,efron1993bootstrap}. Other directional differences
are reported with confidence intervals only. Contextual baselines, edit variants, and SafeDrug were outside the confirmatory
inferential family and are reported descriptively as point estimates.

\section{Results}
\label{sec:results}

\subsection{Main comparison} 
R-GEAN attains the highest edit composite, the predefined summary of addition,
removal, changed-regimen, and action-pattern performance. Its score of 0.4643
exceeds the shared neural model at 0.4350 and RETAIN-FullSet at 0.4275, the
strongest complete-set comparator.

MICRON obtains an edit composite of 0.2884, with addition F1 of 0.2280 and removal F1 of 0.3456. The direct edit variants of RETAIN and GAMENet also underperform their FullSet versions (0.3492 vs.\ 0.4275 and 0.3414 vs.\ 0.4195, respectively). These results indicate that directly predicting additions and removals alone
does not explain R-GEAN's performance. The component analyses below examine the contributions of the asymmetric architecture and candidate-specific inputs.

\begin{table*}[t]
\centering
\setlength{\tabcolsep}{5pt}
\begin{tabular}{lccccc}
\toprule
\textbf{Model} & \textbf{Add F1}$\uparrow$ & \textbf{Remove F1}$\uparrow$
& \textbf{Changed Jacc.}$\uparrow$ & \textbf{Action mF1}$\uparrow$
& \textbf{Composite}$\uparrow$ \\
\midrule
\multicolumn{6}{l}{\emph{Contextual (non-learned) baselines}} \\
\textsc{continuation} & 0.0000 & 0.0000 & \textbf{0.6347} & 0.0467 & 0.1634 \\
\textsc{frequency} & 0.1942 & 0.5014 & 0.5968 & 0.3364 & 0.4012 \\
\midrule
\multicolumn{6}{l}{\emph{Public models, reimplemented under this benchmark}} \\
RETAIN-FullSet & \textbf{0.2687} & 0.4671 & 0.6209 & 0.3811 & 0.4275 \\
RETAIN-Edit & 0.0959 & 0.4780 & 0.5444 & 0.3608 & 0.3492 \\
GAMENet-FullSet & 0.2441 & 0.4658 & \underline{0.6271} & 0.3757 & 0.4195 \\
GAMENet-Edit & 0.0872 & 0.4794 & 0.5202 & 0.3702 & 0.3414 \\
MICRON-Adapted & 0.2280 & 0.3456 & 0.3122 & 0.2684 & 0.2884 \\
\midrule
\multicolumn{6}{l}{\emph{Internal learned baselines}} \\
\textsc{tree} & 0.1955 & 0.5482 & 0.5882 & \underline{0.4124} & 0.4212 \\
\textsc{shared} & 0.1946 & \underline{0.5941} & 0.5919 & 0.4068 & \underline{0.4350} \\
\midrule
\textbf{R-GEAN} & \underline{0.2509} & \textbf{0.6013} & 0.6088 & \textbf{0.4392} & \textbf{0.4643} \\
\bottomrule
\end{tabular}
\caption{Test results for the primary seed (49,032 admissions; 16,444
patients). Bold and underlined values indicate the best and second-best result
in each column. The continuation baseline attains the highest changed-admission
Jaccard despite predicting no additions or removals. The composite is the
predefined summary of edit performance described in
Equation~\ref{eq:composite}.}
\label{tab:main}
\end{table*}

\subsection{Where the advantage lies}

Patient-level paired bootstrap analysis supports R-GEAN's edit-composite
advantage over all four learned comparators. R-GEAN improves the composite by
0.0368 over RETAIN-FullSet, 0.0448 over GAMENet-FullSet, 0.0431 over the tree
model, and 0.0293 over the primary shared baseline. Its advantage over MICRON
is 0.1759. Reported differences are computed from unrounded estimates and may
differ by up to 0.0001 from differences between the rounded table entries.

Its largest advantage is in correctly identifying medication classes active at
24 hours but absent at discharge. Compared with MICRON, R-GEAN improves removal
F1 by 0.2556. It also improves removal F1 over RETAIN-FullSet and
GAMENet-FullSet by 0.1342 and 0.1355, respectively, with 95\% confidence
intervals entirely above zero. R-GEAN additionally exceeds MICRON in
resulting-set micro-F1 by 0.3131.

All eleven 95\% confidence intervals exclude zero, and all corresponding
comparisons remain significant after joint Holm correction. No bootstrap
replicate favors a comparator. Full confidence intervals and empirical
$p$-value calculations are provided in the supplement.

Addition prediction is the only direction in which a complete-set model
outperforms R-GEAN. RETAIN-FullSet achieves higher addition F1
($-0.0178$ relative to R-GEAN; 95\% CI
$[-0.0211,-0.0145]$). R-GEAN nevertheless exceeds GAMENet-FullSet
($+0.0067$, 95\% CI $[0.0035,0.0099]$) and both internal learned baselines.
These directional comparisons were not included in the Holm-adjusted primary analysis and are reported with confidence intervals only.

\subsection{Robustness to composite weighting}

R-GEAN remains the highest-ranked model under all six predefined alternative
weighting schemes and under 94.1\% of 20,000 randomly sampled weight
combinations. It ranks within the top three in 99.99\% of these combinations.

\subsection{Set-level and edit-level rankings disagree}

Models receive different rankings when evaluated on the complete reconstructed
discharge regimen and when evaluated directly on additions and removals. The
rank correlation between resulting-set micro-F1 and the edit composite is weak
($\rho=0.20$, $n=10$).

The continuation baseline ranks first on resulting-set micro-F1 (0.8169)
despite predicting no additions or removals. Similarly, GAMENet-FullSet has the
highest resulting-set micro-F1 among the learned models (0.7938) but ranks fifth
on the edit composite. These results support reporting both evaluation views:
complete-set metrics measure reconstruction of the full discharge regimen,
whereas edit-level metrics measure recovery of medication changes.

\section{Diagnostics and Limitations}

\paragraph{Component attribution.}
The ablations show that addition and removal prediction depend on different
information. The interaction regularizer and prior-history features contribute
most to addition performance, whereas medication-specific exposure contributes
most to removal performance. Removing the interaction regularizer reduces the
composite by 0.0409, and removing candidate exposure from the removal branch
reduces removal F1 by 0.0772 and the composite by 0.0293.
Table~\ref{tab:ablation} summarizes the main component effects.

The shared-model experiments further support the use of separate predictors. The primary shared baseline achieves a composite of 0.4350. A separately retrained shared-predictor ablation achieves 0.4345 ($\Delta S=-0.0298$). After widening the shared model to match R-GEAN's parameter count, the parameter-matched control remains 0.0225 behind R-GEAN at seed 2026 (95\% CI $[0.0208,0.0242]$).

\begin{table}[t]
\centering
{\small
\setlength{\tabcolsep}{2pt}
\begin{tabular}{@{}lrrr@{}}
\toprule
\textbf{Variant}
  & $\Delta\mathrm{F1}^{+}$
  & $\Delta\mathrm{F1}^{-}$
  & $\Delta S$ \\
\midrule
Add.\ w/o DDI        & $-0.0786$ & ---       & $-0.0409$ \\
Add.\ w/o history    & $-0.0299$ & ---       & $-0.0224$ \\
Add.\ w/o transition & $-0.0125$ & ---       & $-0.0108$ \\
Add.\ w/o exposure   & $-0.0227$ & ---       & $-0.0085$ \\
Rem.\ w/o exposure   & ---       & $-0.0772$ & $-0.0293$ \\
Shared predictor (rerun)     & $-0.0570$ & $-0.0077$ & $-0.0298$ \\
\bottomrule
\end{tabular}
}
\caption{Changes in test performance after removing individual components.
Full R-GEAN obtains $\mathrm{F1}^{+}=0.2509$,
$\mathrm{F1}^{-}=0.6013$, and $S=0.4643$. When one branch is modified, the
other is held fixed. Complete results are provided in the supplement.}
\label{tab:ablation}
\end{table}

\paragraph{Error structure.}
R-GEAN's clearest weakness is ranking rare additions. Across the 39 classes in
the bottom half of training-set addition support, R-GEAN attains an addition F1
of 0.0004, the lowest among the evaluated models. All of these classes are
available to the model under Equation~\ref{eq:candidates}, but the correct
classes usually receive low scores. Rare-class recall@5 is 0.017, recall@10 is 0.029, and mean reciprocal rank is
0.042; recall@5 is zero in the lowest-support quartile. Validation-selected
focal, class-balanced, and logit-adjusted losses do not produce a stable
improvement.

Multi-edit admissions, which contain several simultaneous medication changes,
also remain difficult. Among 13{,}560 such admissions, R-GEAN exactly recovers
all changes in 0.0016 of cases. It nevertheless achieves the strongest partial recovery, with addition recall
of 0.256 and removal recall of 0.626.

Removal probabilities are substantially less well calibrated than addition probabilities. Removal ECE is 0.1943, compared with an addition ECE of 0.0323, and the removal Brier score is 0.1370 \citep{guo2017calibration,naeini2015calibration,brier1950verification}. 
Isotonic calibration fitted on the validation set reduces removal ECE to
0.0023 while leaving removal F1 nearly unchanged at 0.6011. Calibration
therefore improves probability estimates rather than change recovery, and the
raw probabilities should not be interpreted as clinical confidence.

\paragraph{Removal-label composition.}
R-GEAN's removal advantage is not limited to medications given only briefly.
True removals consist mainly of anti-infectives (38.2\%) and cardiovascular
agents (20.9\%), and 26.1\% of medication orders active at 24 hours last no more
than one day. In a post-hoc analysis, removal classes are grouped by their
median order duration in the test set. This grouping is not used for training,
threshold selection, or model selection. Relative to RETAIN-FullSet, R-GEAN
improves removal F1 for both classes with median duration
$\leq48$ hours ($+0.099$) and longer-duration classes ($+0.199$). Some removal
labels still reflect the natural completion of short medication courses.

\paragraph{Interaction burden.}
Pair-normalized interaction burden is similar across models
(0.968--0.996; R-GEAN 0.9948), providing no evidence of a safety advantage.
Removing the interaction term leads to overprediction, with 3.9 additions per
admission and addition F1 of 0.172. In matched diagnostic reruns, a
training-derived count-matching regularizer achieves nearly the same composite
as the interaction-regularized model (0.4615 vs.\ 0.4618), without using
interaction information. The TWOSIDES penalty therefore mainly limits the
number of predicted additions; it does not provide evidence of safer
medication combinations.

\paragraph{Robustness and reproducibility.}
R-GEAN performs broadly similarly across most age, sex, and admission-type
strata, although several small admission-type groups show greater variation. These subgroup
results are internal stress tests and should not be interpreted as a
comprehensive fairness or external-validity analysis. Comparisons with public
models should be interpreted at the whole-system level because the models
differ in task adaptation and available inputs. The matched shared model
provides the more controlled evidence for separate addition and removal
predictors.

An end-to-end rerun of the data-processing and evaluation pipeline reproduced
the benchmark and reported results. R-GEAN has 465{,}924 trainable parameters,
and training takes 35.7\,s on a single RTX 4090. Further computational and
reproducibility details are provided in the supplement.

\paragraph{Scope and limitations.}
The benchmark captures net medication-class changes between 24 hours after
admission and discharge. It does not represent a reconciled discharge
medication list, every medication event during the admission, dose or route
changes, within-class substitutions, or treatment appropriateness. Only
27.7\% of 708{,}149 medication start and stop events alter membership between
the two snapshots.

The cohort is drawn from a single health system and includes only patients with
prior admissions. Generalization to first admissions, later time periods, and
other institutions remains untested. The prescription data used here do not provide reliable explicit fields for
PRN status, cancellations, or discontinuation events; these cases are
handled using prescription intervals and medication-mapping rules.

The study did not include clinician review, external or temporal validation, or
a comprehensive fairness analysis.

\section{Conclusion}

R-GEAN predicts which ATC3 medication classes are added to or removed from the
regimen active at 24 hours by discharge. It achieves the highest edit composite
among the evaluated models, with its advantage concentrated in removal
prediction. Rare additions and multi-edit cases remain difficult. Complete-set
and edit-level metrics also produce different model rankings, supporting the
use of both evaluation views. These retrospective results describe observed prescribing in a single-system
cohort restricted to patients with prior admissions.

\section*{Ethical Statement}

This benchmark is designed to test whether medication-prediction models recover
observed medication changes or mainly copy therapies that remain unchanged. It
is intended for research evaluation, model comparison, and failure analysis,
not for medication recommendation or clinical decision support.

MIMIC-IV is de-identified and was accessed under PhysioNet credentialing and its data use agreement. Code and configurations will be released for research use; the data must be obtained separately through PhysioNet.

Because the targets reflect historical prescribing, they may encode institutional practices and disparities and do not establish clinical appropriateness, efficacy, or safety. R-GEAN has not been clinically validated and must not be used for prescribing or patient-facing decisions.

\bibliography{references_consistency_hardened}

\clearpage
\appendix
\setcounter{secnumdepth}{2}

\section*{Supplementary Material}



\section{Benchmark construction and validity}

\subsection{Mapping and cohort summary}

Prescription strings are normalized through RxNorm, which unifies names and formulations of the same drug
concept, and mapped to the Anatomical Therapeutic Chemical (ATC) system; ingredients with several candidate
codes are resolved to the route- and anatomy-compatible ATC branch. After applying the prespecified exclusion policy, records are mapped to the
78 ATC3 classes used for prediction. The prediction vocabulary $\mathcal{V}$ comprises 78 ATC3 therapeutic classes
defined by a prespecified clinical-inclusion policy, all represented in the
training cohort.

\begin{table}[!ht]
\centering
\small
\setlength{\tabcolsep}{6pt}
\begin{tabular}{@{}lr@{}}
\toprule
\textbf{Stage} & \textbf{Result} \\
\midrule
Raw prescription rows                       & 20{,}292{,}611 \\
Records linked to eligible admissions       & 15{,}562{,}624 \\
RxNorm entity coverage                      & 64.6\% \\
ATC coverage (by record)                    & 88.6\% \\
Resolved eligible records (after exclusions)& 99.4\% \\
Final ATC3 vocabulary                        & 78 classes \\
Patients                                     & 82{,}286 \\
Admissions                                   & 240{,}480 \\
\bottomrule
\end{tabular}
\caption{Summary of medication mapping and benchmark construction. Coverage
percentages use the denominators defined in the text.}
\label{tab:mapping}
\end{table}

The three coverage percentages use different denominators. RxNorm entity coverage
is the share of the 34{,}000 unique medication entities---keyed by normalized
name, national drug code, generic sequence number, formulary code, product
strength, and route---resolved through the national-drug-code-to-RxNorm and
RxNorm-consensus crosswalks (21{,}979 of 34{,}000). ATC coverage by record is the
record-weighted share of the 15{,}562{,}624 linked prescription records that carry
an ATC code (88.6\%). Resolved eligible records (99.4\%) is the record-weighted
share that are either mapped or routed to a documented non-regimen exclusion by
the frozen policy, computed after policy exclusions. Drug strings are normalized
by trimming, upper-casing, and collapsing internal whitespace, with no stemming or
synonym substitution. Entities with several candidate ATC codes are resolved by a route-gated
multi-ATC correction to the route- and anatomy-compatible branch. Long-tail
entities that remain unresolved are left unmapped, with none exceeding a
0.1\% record-frequency threshold. Mapped medications are collapsed to
per-admission ATC3 membership. The frozen prediction vocabulary contains
78 ATC3 therapeutic classes selected under the prespecified
clinical-inclusion policy. All 78 classes are represented in the training
cohort.

\subsection{Anchor and target construction}

The anchor regimen contains the ATC3 medication classes recorded as active
24 hours after admission. The target regimen contains the classes recorded as
active at discharge, as determined from prescription start and stop times.
Because MIMIC-IV does not provide a reconciled discharge medication list, the
target regimen is an interval-based approximation of the medication classes
active at discharge.

Multiple prescription records mapped to the same ATC3 class are collapsed to
set membership at each landmark. Dose and route changes do not create separate
targets, and substitutions within the same ATC3 class are not represented.
Temporary exposures and stop--restart sequences that do not change membership
between the two landmarks are not included in the net addition and removal
labels.

\subsection{Patient-disjoint split and cohort selection}

Patients are assigned to train, validation, and test sets in a
$70{:}10{:}20$ ratio using a seeded random permutation of patient identifiers
(seed 2026). Assignment is not stratified by calendar year, and all admissions
from a patient remain in one split. No patient appears in more than one split.

Eligible admissions require at least one completed prior admission so that
patient history is available. The 24-hour regimen is constructed only from the
current admission. Compared with otherwise eligible single-admission adults,
the retained cohort is older and has longer hospital stays
(Table~\ref{tab:cohort}).

\begin{table}[!tbp]
\centering
\small
\setlength{\tabcolsep}{6pt}
\begin{tabular}{@{}lrrr@{}}
\toprule
\textbf{Variable} & \textbf{Recurrent-care} & \textbf{Single-adm.} & \textbf{SMD} \\
\midrule
Admissions            & 240{,}480 & 109{,}899 & --- \\
Mean age (years)      & 62.9 & 56.6 & 0.33 \\
Mean length of stay (d)    & 6.1 & 5.2 & 0.12 \\
\bottomrule
\end{tabular}
\caption{Cohort selection relative to otherwise-eligible single-admission adults (standardized mean
differences).}
\label{tab:cohort}
\end{table}

\subsection{Leakage and feature provenance}

Current-admission clinical features are gated by their \emph{availability}
time rather than event time. An observation is included only if its recording
time is at or before the 24-hour anchor; values measured within the window but
recorded later are excluded. An order is marked as discontinued only when its
stop time is at or before the anchor. Table~\ref{tab:prov} groups the input blocks by their availability rule; all
active blocks pass the automated leakage checks.

\begin{table*}[!t]
\centering
\footnotesize
\setlength{\tabcolsep}{6pt}
\begin{tabular}{@{}p{4.2cm}p{7.0cm}p{1.8cm}c@{}}
\toprule
\textbf{Input block} &
\textbf{Availability rule} &
\textbf{Branch} &
\textbf{Verdict} \\
\midrule
Prior-admission and admission features
& Available by admission
& Add, Rem
& Pass \\

Anchor regimen (multi-hot)
& Active at the 24-hour landmark
& Add, Rem
& Pass \\

Laboratory summaries and candidate-specific interactions
& Recorded at or before the 24-hour cutoff
& Add, Rem
& Pass \\

Laboratory and vital-sign transition states
& Recorded at or before the 24-hour cutoff
& Add
& Pass \\

Candidate-specific medication exposure
& Order information available by the 24-hour cutoff
& Add, Rem
& Pass \\

Candidate metadata
& Derived from the training split only
& Add, Rem
& Pass \\
\bottomrule
\end{tabular}
\caption{Feature provenance, availability rules, branch usage, and automated
leakage-check results.}
\label{tab:prov}
\end{table*}

The addition loss also uses an external TWOSIDES interaction matrix as a
training-time regularizer. The matrix is split-independent and is not a
patient-level model input.

The cohort is defined from the HOSP module. Laboratory measurements are
obtained from \texttt{hosp/labevents}, whereas vital signs are obtained from
linked \texttt{icu/chartevents} records. In the primary run, seven vital
variables received train-fitted transition bands and entered R-GEAN through the
addition-branch transition representation; \texttt{FiO2} did not meet the
minimum finite-observation requirement. Vital-sign coverage was approximately
13\% of admissions, and unavailable vital-sign information was represented by
no-data transition states.

\subsection{Action-stratum definitions}

Let the anchor and target sizes be the numbers of classes active at the
24-hour landmark and at discharge, respectively, and let the addition and
removal counts be $\mathrm{add}=|Y^{+}|$ and
$\mathrm{rem}=|Y^{-}|$. Each admission
is assigned to one of seven strata by the conditions in Table~\ref{tab:strata}, evaluated in the predefined
first-match order shown. The categories form an exhaustive, mutually exclusive partition under this precedence
order.

\begin{table}[!tbp]
\centering
\small
\setlength{\tabcolsep}{6pt}
\begin{tabular}{@{}rlp{4.7cm}@{}}
\toprule
\textbf{\#} & \textbf{Stratum} & \textbf{Condition} \\
\midrule
1 & Empty-to-nonempty & anchor size $=0$, target size $>0$ \\
2 & Nonempty-to-empty & anchor size $>0$, target size $=0$ \\
3 & Continue    & $\mathrm{add}=0$ and $\mathrm{rem}=0$ \\
4 & Add-only    & $\mathrm{add}>0$ and $\mathrm{rem}=0$ \\
5 & Remove-only & $\mathrm{rem}>0$ and $\mathrm{add}=0$ \\
6 & Switch      & $\mathrm{add},\mathrm{rem}>0$ and $\mathrm{add}+\mathrm{rem}\leq 3$ \\
7 & Multi-edit  & $\mathrm{add},\mathrm{rem}>0$ and $\mathrm{add}+\mathrm{rem}>3$ \\
\bottomrule
\end{tabular}
\caption{Action strata and their first-match precedence order.}
\label{tab:strata}
\end{table}

The three-edit boundary separating switch from multi-edit admissions and the
first-match precedence order were fixed before test evaluation.

\section{Baseline and statistical details}

\subsection{Baseline implementations}

RETAIN and GAMENet are evaluated in FullSet and Edit forms as described in the
main paper. For the within-admission adaptation, MICRON receives
current-admission diagnosis and procedure bags and the observed 24-hour regimen
as its previous medication state. Its previous-state clinical input is empty,
and predictions are not propagated across admissions. The shared model receives
all R-GEAN feature blocks plus a general clinical summary. SafeDrug is adapted to ATC3 and reported as a sensitivity analysis, obtaining
an edit composite of 0.2816. It is excluded from the primary comparison and
rank-correlation analysis.

The Tree baseline is a LightGBM candidate-pair classifier with 63 leaves,
learning rate 0.05, 50 minimum samples per leaf, feature and bagging fractions
of 0.8, and 300 estimators per direction. It uses a 3:1 negative-to-positive
training ratio and validation-selected thresholds.

\subsection{Bootstrap confidence intervals}

Uncertainty uses a patient-level paired bootstrap: patients are sampled with
replacement, all admissions of each sampled patient are retained, and both
compared models are evaluated on the identical resampled admissions so that
differences are paired. Each replicate recomputes the pooled metric from scratch
($B=10{,}000$), and confidence intervals are the 2.5th and 97.5th percentiles of
the replicate differences. The two-sided empirical $p$-value uses the add-one
estimator $2\,(1+\min(b^{-},b^{+}))/(B+1)$, where $b^{-}$ and $b^{+}$ count
replicates below and above zero; at $B=10{,}000$ its minimum attainable value is
approximately $0.0002$. Holm correction is applied jointly across the eleven
comparisons in Table~\ref{tab:holm}; because every comparison attains the
empirical floor, the common Holm-adjusted value is $0.0022$, and all eleven are
rejected. Exact-set accuracy, which appears in the table, is the fraction of
admissions whose reconstructed discharge regimen exactly matches the observed
discharge regimen.

\begin{table}[!tbp]
\centering
\footnotesize
\setlength{\tabcolsep}{4pt}
\begin{tabular}{@{}llrr@{}}
\toprule
\textbf{Comparator} & \textbf{Metric} & \textbf{Diff.} & \textbf{95\% CI} \\
\midrule
RETAIN-FullSet  & composite & 0.0368 & [0.0341, 0.0395] \\
GAMENet-FullSet & composite & 0.0448 & [0.0424, 0.0472] \\
Tree            & composite & 0.0431 & [0.0408, 0.0456] \\
Shared          & composite & 0.0293 & [0.0276, 0.0310] \\
MICRON & composite            & 0.1759 & [0.1731, 0.1788] \\
MICRON & addition F1          & 0.0229 & [0.0193, 0.0264] \\
MICRON & removal F1           & 0.2556 & [0.2509, 0.2604] \\
MICRON & changed Jaccard      & 0.2966 & [0.2924, 0.3006] \\
MICRON & action macro-F1      & 0.1708 & [0.1633, 0.1816] \\
MICRON & resulting-set micro-F1 & 0.3131 & [0.3088, 0.3173] \\
MICRON & exact-set accuracy   & 0.1066 & [0.1033, 0.1100] \\
\bottomrule
\end{tabular}
\caption{R-GEAN improvements over each comparator with 95\% patient-paired bootstrap confidence intervals.
Positive values favor R-GEAN.}
\label{tab:holm}
\end{table}

\subsection{Composite-weight sensitivity}

R-GEAN ranks first under all six prespecified weighting schemes
(Table~\ref{tab:weights}) and under the unweighted component mean. Over 20{,}000
random simplex weightings it ranks first in 94.1\% of settings and within the top
three in 99.99\% (worst rank four); within the edit-centric region (addition-F1
$+$ removal-F1 weight $\geq 0.5$; 9{,}981 weightings) it ranks first in 94.0\% and
within the top three in 100\%. R-GEAN is also Pareto-optimal over the four score
components. Random weightings are drawn from a uniform Dirichlet distribution over
the four components (seed 2026, 20{,}000 samples over all eleven systems); ties
are broken by the minimum rank; and a system is Pareto-optimal if no other system
is at least as high on all four components and strictly higher on one.

\begin{table}[!tbp]
\centering
\small
\setlength{\tabcolsep}{6pt}
\begin{tabular}{@{}lrrrr@{}}
\toprule
\textbf{Scheme} & \textbf{F1$^{+}$} & \textbf{F1$^{-}$} & \textbf{$J_{\text{chg}}$} & \textbf{F1$_{\text{act}}$} \\
\midrule
Primary        & 0.35 & 0.30 & 0.25 & 0.10 \\
Equal          & 0.25 & 0.25 & 0.25 & 0.25 \\
Edit-heavy     & 0.40 & 0.40 & 0.10 & 0.10 \\
Balanced       & 0.30 & 0.30 & 0.20 & 0.20 \\
Addition-heavy & 0.45 & 0.25 & 0.20 & 0.10 \\
Removal-heavy  & 0.25 & 0.45 & 0.20 & 0.10 \\
\bottomrule
\end{tabular}
\caption{Prespecified composite-weight schemes over addition F1, removal F1,
changed-admission Jaccard, and action macro-F1.}
\label{tab:weights}
\end{table}

\subsection{MICRON, DualNN, and R-GEAN comparison}

Table~\ref{tab:novelty} compares the prediction settings and model structures
of the adapted MICRON baseline, DualNN, and R-GEAN. Under the shared
within-admission setting, R-GEAN differs from the adapted MICRON through its
disjoint candidate spaces and candidate-specific exposure.

\begin{table}[!tbp]
\centering
\footnotesize
\setlength{\tabcolsep}{4pt}
\begin{tabular}{@{}p{1.7cm}p{1.7cm}p{1.5cm}p{1.9cm}@{}}
\toprule
\textbf{Property} & \textbf{MICRON} & \textbf{DualNN} & \textbf{R-GEAN} \\
\midrule
Prediction interval & within admission & between visits & within admission \\
Base regimen & observed 24-hour anchor & predicted & observed 24-hour anchor \\
Recursive prediction & no & no & no \\
Candidate spaces & one shared vector & two nets, shared space & two disjoint anchor spaces \\
Candidate-specific exposure & no & no & yes \\
Observation boundary & 24-hour cutoff & visit-level input & explicit 24-hour cutoff \\
Dataset & MIMIC-IV & MIMIC-III & MIMIC-IV \\
\bottomrule
\end{tabular}
\caption{Comparison of the adapted MICRON baseline, DualNN, and R-GEAN under
the within-admission prediction setting. MICRON columns describe the adapted
implementation evaluated here; DualNN is listed as originally proposed and was
not re-evaluated.}
\label{tab:novelty}
\end{table}

\section{Robustness and Reproducibility}

\subsection{Multiseed results}
\label{sec:multiseed}

Each model is trained independently under seeds 2026--2028, and predictions
are not pooled. Table~\ref{tab:multiseed} reports the exact seed-specific edit
composites and their arithmetic means. R-GEAN achieves the highest edit
composite under all three seeds.

For the multiseed robustness analysis, R-GEAN uses the thresholds fixed under
the original multiseed protocol, $\tau^{+}=0.30$ and $\tau^{-}=0.65$, across
all three seeds. These differ from the validation-reselected primary
thresholds, $\tau^{+}=0.25$ and $\tau^{-}=0.65$, used for the headline test
result. Comparison models select thresholds from their corresponding
validation predictions. We therefore interpret the multiseed analysis as a
ranking robustness check rather than a fully protocol-matched threshold
comparison.

\begin{table}[!tbp]
\centering
\footnotesize
\setlength{\tabcolsep}{4pt}
\begin{tabular*}{\linewidth}{
@{\extracolsep{\fill}}
l
S[table-format=1.4]
S[table-format=1.4]
S[table-format=1.4]
S[table-format=1.4]
@{}
}
\toprule
\textbf{Model}
& \multicolumn{1}{c}{\textbf{2026}}
& \multicolumn{1}{c}{\textbf{2027}}
& \multicolumn{1}{c}{\textbf{2028}}
& \multicolumn{1}{c}{\textbf{Mean}} \\
\midrule
R-GEAN          & 0.4672 & 0.4583 & 0.4655 & 0.4636 \\
Shared          & 0.4350 & 0.4278 & 0.4375 & 0.4334 \\
RETAIN-FullSet  & 0.4275 & 0.4292 & 0.4289 & 0.4285 \\
GAMENet-FullSet & 0.4195 & 0.4170 & 0.4182 & 0.4182 \\
MICRON-Adapted  & 0.2884 & 0.2895 & 0.2894 & 0.2891 \\
\bottomrule
\end{tabular*}
\caption{Seed-specific edit composites and arithmetic means. Means are
calculated from the full-precision seed-specific results.}
\label{tab:multiseed}
\end{table}

\subsection{Parameter-matched shared control}

We widen the shared model to approximately match R-GEAN's parameter count
without consulting test results. Across three seeds, the matched shared model
improves over the original shared model by 0.0029 but remains 0.0274 below
R-GEAN on average (Table~\ref{tab:parammatched}). At seed 2026, the paired
difference is 0.0225 (95\% CI $[0.0208,0.0242]$). This control supports
direction-specific predictors beyond parameter count, although it does not
isolate all architectural differences.

\begin{table}[!tbp]
\centering
\small
\setlength{\tabcolsep}{6pt}
\begin{tabular*}{\linewidth}{
@{\extracolsep{\fill}}
l
r
S[table-format=1.4]
@{\,${}\pm{}$\,}
S[table-format=1.4]
@{}
}
\toprule
\textbf{Model}
& \textbf{Parameters}
& \multicolumn{2}{c}{\textbf{Composite}} \\
\midrule
Shared, original      & 363{,}394 & 0.4333 & 0.0050 \\
Shared, param-matched & 465{,}454 & 0.4362 & 0.0070 \\
R-GEAN                & 465{,}924 & 0.4636 & 0.0047 \\
\bottomrule
\end{tabular*}
\caption{Three-seed parameter-matched comparison. Models were retrained under
a common capacity-control protocol, so the shared-model values differ slightly
from the primary multiseed results.}
\label{tab:parammatched}
\end{table}

\subsection{R-GEAN computational footprint}

R-GEAN contains 465{,}924 trainable parameters. For the primary recorded run
on a single NVIDIA RTX 4090 (PyTorch 2.5.1, Python 3.11.8), optimization took
31.3\,s for the addition branch and 4.4\,s for the removal branch, for 35.7\,s in
total (Table~\ref{tab:compute}). These measurements cover branch optimization
only; preprocessing, data loading, validation, and inference are not included.
They describe the implementation footprint and are not intended as a controlled
speed comparison with the baselines.

\begin{table}[!tbp]
\centering
\small
\setlength{\tabcolsep}{6pt}
\begin{tabular}{@{}lrr@{}}
\toprule
\textbf{Component} & \textbf{Parameters} & \textbf{Training time} \\
\midrule
R-GEAN addition branch & 235{,}138 & 31.3\,s \\
R-GEAN removal branch  & 230{,}786 & 4.4\,s \\
R-GEAN combined        & 465{,}924 & 35.7\,s \\
\bottomrule
\end{tabular}
\caption{R-GEAN parameter counts and recorded branch training times for the
primary seed-2026 run.}
\label{tab:compute}
\end{table}

An end-to-end rerun of the data-processing and evaluation pipeline reproduced
the benchmark cohort, the 78-class vocabulary, and the reported primary
evaluation results.

\section{Diagnostic analyses}

The primary comparison and confirmatory family were prespecified for seed 2026.
The analyses in this section are secondary controls and exploratory diagnostics
on the same test split and are not treated as confirmatory.

\subsection{Set-level and edit-level rankings}

Models receive different rankings on the reconstructed discharge regimen and on
the additions and removals directly (Table~\ref{tab:ranking}). The rank
correlation between resulting-set micro-F1 and the edit composite is weak
(Spearman $\rho=0.20$, $n=10$; no ties occur). Spearman correlation is computed
over the ten systems shown; SafeDrug is excluded because it is reported only as a
sensitivity analysis.

\begin{table}[!tbp]
\centering
\footnotesize
\setlength{\tabcolsep}{3pt}
\begin{tabular*}{\linewidth}{
@{\extracolsep{\fill}}
l
S[table-format=1.4]
S[table-format=2.0]
S[table-format=1.4]
S[table-format=2.0]
@{}
}
\toprule
\textbf{Model}
& \multicolumn{1}{c}{\textbf{Set $\mu$F1}}
& \multicolumn{1}{c}{\textbf{Set rank}}
& \multicolumn{1}{c}{\textbf{Comp.}}
& \multicolumn{1}{c}{\textbf{Edit rank}} \\
\midrule
R-GEAN          & 0.7769 & 4  & 0.4643 & 1  \\
Shared          & 0.7334 & 7  & 0.4350 & 2  \\
RETAIN-FullSet  & 0.7878 & 3  & 0.4275 & 3  \\
Tree            & 0.7425 & 6  & 0.4212 & 4  \\
GAMENet-FullSet & 0.7938 & 2  & 0.4195 & 5  \\
Frequency       & 0.7753 & 5  & 0.4012 & 6  \\
RETAIN-Edit     & 0.6164 & 8  & 0.3492 & 7  \\
GAMENet-Edit    & 0.5988 & 9  & 0.3414 & 8  \\
MICRON-Adapted  & 0.4639 & 10 & 0.2884 & 9  \\
Continuation    & 0.8169 & 1  & 0.1634 & 10 \\
\bottomrule
\end{tabular*}
\caption{Set-level and edit-level ranks for the ten systems included in the
rank-correlation analysis. Rank 1 is best.}
\label{tab:ranking}
\end{table}

\subsection{Component ablations}

Table~\ref{tab:supp_ablation} reports branch-specific ablations with the other
branch held fixed. Candidate exposure contributes most to removal performance,
whereas the interaction regularizer and prior history contribute most to
addition performance. Removing the class-level priors slightly increases the
aggregate composite but lowers rare-addition recall@5 from approximately 0.017 to
0.002.

\begin{table*}[!t]
\centering
\small
\setlength{\tabcolsep}{6pt}
\begin{tabular*}{\linewidth}{
@{\extracolsep{\fill}}
l
S[table-format=1.4]
S[table-format=1.4]
S[table-format=1.4]
S[table-format=1.4]
S[table-format=1.4]
S[table-format=+1.4]
@{}
}
\toprule
\textbf{Variant}
& \multicolumn{1}{c}{\textbf{Add F1}}
& \multicolumn{1}{c}{\textbf{Remove F1}}
& \multicolumn{1}{c}{\textbf{Changed Jacc.}}
& \multicolumn{1}{c}{\textbf{Action macro-F1}}
& \multicolumn{1}{c}{\textbf{Composite}}
& \multicolumn{1}{c}{\textbf{$\Delta$ Comp.}} \\
\midrule
Full R-GEAN
& 0.2509 & 0.6013 & 0.6088 & 0.4392 & 0.4643 & 0.0000 \\

Addition, no transition states
& 0.2384 & 0.6013 & 0.5888 & 0.4251 & 0.4535 & -0.0108 \\

Addition, no candidate exposure
& 0.2282 & 0.6013 & 0.6153 & 0.4173 & 0.4558 & -0.0085 \\

Addition, no laboratory interaction
& 0.2441 & 0.6013 & 0.5966 & 0.4322 & 0.4582 & -0.0061 \\

Addition, no prior history
& 0.2210 & 0.6013 & 0.5837 & 0.3825 & 0.4419 & -0.0224 \\

Addition, no interaction regularizer
& 0.1723 & 0.6013 & 0.5546 & 0.4410 & 0.4234 & -0.0409 \\

Removal, no candidate exposure
& 0.2509 & 0.5241 & 0.5980 & 0.4047 & 0.4350 & -0.0293 \\

Removal, no laboratory interaction
& 0.2509 & 0.5956 & 0.6099 & 0.4379 & 0.4627 & -0.0016 \\

Removal, no prior history
& 0.2509 & 0.5978 & 0.6117 & 0.4327 & 0.4633 & -0.0010 \\

Shared predictor
& 0.1939 & 0.5936 & 0.5882 & 0.4149 & 0.4345 & -0.0298 \\

No class-level priors
& 0.2508 & 0.6002 & 0.6185 & 0.4404 & 0.4665 & +0.0022 \\
\bottomrule
\end{tabular*}
\caption{Complete component ablations on the test set. The unmodified
prediction direction is held fixed.}
\label{tab:supp_ablation}
\end{table*}

The final model retains the prespecified class-level priors because they improve
rare-addition ranking, although removing them slightly increases the aggregate
composite. This secondary ablation was conducted after the primary architecture
and test evaluation had been locked.

\subsection{Cardinality and interaction regularization}

The interaction-derived regularizer mainly controls addition over-prediction. Table~\ref{tab:cardinality}
compares it with generic cardinality controls: a train-derived count-matching regularizer, using no interaction
structure, reproduces the predictive effect of the interaction-derived term, and the pair-normalized interaction
burden does not fall when the term is added. We therefore read the term as over-prediction (cardinality) control
rather than interaction-specific predictive information.

The interaction term $\Omega_{\mathrm{ddi}}$ is a symmetric $78\times78$
class-level matrix derived from the external TWOSIDES resource, remapped from
drug ingredients to the 78 ATC3 classes, with zero diagonal and 2{,}918
unordered interacting class pairs, corresponding to 5{,}836 nonzero
off-diagonal matrix entries. The addition penalty is
$\sum_{i,j} a_i\,\Omega_{ij}\,R_j$, where $a$ is the vector of predicted
addition probabilities and $R=\min(1,\,M_A+a)$ is the addition-adjusted regimen
formed from the anchor regimen $M_A$ and the predicted additions. The penalty
therefore covers interactions between predicted additions and both the
already-present anchor medications and other predicted additions. The
pair-normalized interaction burden is the expected number of interacting pairs in
the predicted discharge regimen divided by the number of unordered regimen pairs.
The L1 expected-count control penalizes the mean predicted addition count; the
anchor-size count-matching control penalizes deviation from a train-derived target
count $k_i$ equal to the mean number of true additions in the admission's
anchor-size bin, using no target or action information.

\begin{table}[!tbp]
\centering
\small
\setlength{\tabcolsep}{5pt}
\begin{tabular}{@{}lrrr@{}}
\toprule
\textbf{Variant} & \textbf{Comp.} & \textbf{Add F1} & \textbf{Add/adm} \\
\midrule
No regularizer                & 0.4223 & 0.1686 & 3.69 \\
Interaction-derived regularizer & 0.4618 & 0.2454 & 1.95 \\
L1 expected-count control     & 0.4467 & 0.2084 & 2.77 \\
Anchor-size count-matching    & 0.4615 & 0.2448 & 1.87 \\
\bottomrule
\end{tabular}
\caption{Addition-branch regularization (removal branch fixed): a count-matching control reproduces the
predictive effect of the interaction-derived term. Pair-normalized interaction burden is 0.983 without the term
and 0.994 with it.}
\label{tab:cardinality}
\end{table}

\subsection{Rare additions and calibration}

Across the five models with stored rare-class F1 results, performance is low
for rare additions, and R-GEAN attains the lowest rare-addition F1
(Table~\ref{tab:rare_f1}). Rare additions are also poorly ranked before
thresholding: recall@5 is 0.017 for rare classes and zero in the
lowest-support quartile (Table~\ref{tab:rare}). Recall@$k$ is the mean over
true additions of the indicator that the class ranks within the top $k$
candidates; MRR is the mean reciprocal rank; and macro-AUPRC is the unweighted
mean of per-class average precision over candidate positions. Classes with no
test additions contribute no ranking events and are excluded from macro-AUPRC.
Across the validation comparisons, focal, class-balanced, and logit-adjusted
losses did not yield a stable improvement. In the matched test rerun, the
validation-selected logit-adjusted variant leaves rare-class recall@5 at zero
and lowers the composite from 0.4629 to 0.4593.

\begin{table}[!tbp]
\centering
\small
\begin{tabular}{@{}lr@{}}
\toprule
\textbf{Model} & \textbf{Rare-addition F1} \\
\midrule
Tree             & 0.0938 \\
RETAIN-FullSet   & 0.0765 \\
GAMENet-FullSet  & 0.0541 \\
Shared           & 0.0231 \\
R-GEAN           & 0.0004 \\
\bottomrule
\end{tabular}
\caption{Addition F1 across the 39 classes in the bottom half of training-set
addition support.}
\label{tab:rare_f1}
\end{table}

Validation-only isotonic calibration substantially reduces removal ECE while
leaving F1 nearly unchanged (Table~\ref{tab:calib}). Calibration uses 15
equal-width probability bins; expected calibration error is computed at the
candidate level, separately for the addition and removal directions. Isotonic
calibrators are fit on the validation candidate predictions only, one per
direction, and applied once to the fixed test predictions; the reported F1 uses a
threshold re-selected on validation after calibration, and the Brier score is
averaged over candidate cells. Calibration therefore improves probability quality
rather than edit recovery; raw probabilities should not be interpreted as clinical
confidence.

\begin{table}[!tbp]
\centering
\small
\setlength{\tabcolsep}{6pt}
\begin{tabular}{@{}lrrrr@{}}
\toprule
\textbf{Group} & \textbf{R@5} & \textbf{R@10} & \textbf{MRR} & \textbf{Macro-AUPRC} \\
\midrule
All classes        & 0.384 & 0.550 & 0.268 & 0.076 \\
Rare classes       & 0.017 & 0.029 & 0.042 & 0.028 \\
Common classes     & 0.440 & 0.630 & 0.302 & 0.123 \\
Rarest quartile    & 0.000 & 0.000 & 0.019 & 0.007 \\
\bottomrule
\end{tabular}
\caption{Addition ranking quality by class-frequency group.}
\label{tab:rare}
\end{table}

\begin{table}[!tbp]
\centering
\small
\setlength{\tabcolsep}{6pt}
\begin{tabular}{@{}llrrr@{}}
\toprule
\textbf{Direction} & \textbf{Method} & \textbf{ECE} & \textbf{Brier} & \textbf{F1} \\
\midrule
Addition & raw      & 0.0323 & 0.0235 & 0.2509 \\
Addition & isotonic & 0.0003 & 0.0213 & 0.2510 \\
Removal  & raw      & 0.1943 & 0.1370 & 0.6013 \\
Removal  & isotonic & 0.0023 & 0.0904 & 0.6011 \\
\bottomrule
\end{tabular}
\caption{Validation-only calibration. Calibration improves probability quality with essentially unchanged
thresholded F1.}
\label{tab:calib}
\end{table}

\subsection{Subgroup stress tests}

Table~\ref{tab:subgroup} reports performance across age band, sex, and
admission type. Performance is broadly similar across most age, sex, and admission-type
strata, although several small admission-type groups show greater variation. Age bands and
admission-type categories follow the recorded MIMIC-IV fields, and each
admission contributes to one subgroup per dimension. These results are
descriptive internal stress tests and do not constitute a comprehensive
fairness or external-validity analysis.

\begin{table*}[!t]
\centering
\footnotesize
\setlength{\tabcolsep}{5pt}
\begin{tabular}{@{}lrrrrr@{}}
\toprule
\textbf{Subgroup} & \textbf{Adm.} & \textbf{Add F1} & \textbf{Rem F1} & \textbf{Chg. Jacc.} & \textbf{Comp.} \\
\midrule
Age 18--44 & 8{,}254  & 0.2378 & 0.5982 & 0.6197 & 0.4613 \\
Age 45--64 & 16{,}171 & 0.2361 & 0.6064 & 0.6133 & 0.4599 \\
Age 65--74 & 10{,}865 & 0.2530 & 0.6065 & 0.5955 & 0.4624 \\
Age 75--84 & 8{,}292  & 0.2509 & 0.5913 & 0.5875 & 0.4552 \\
Age 85+    & 5{,}450  & 0.2414 & 0.5930 & 0.5698 & 0.4476 \\
Sex F      & 25{,}473 & 0.2467 & 0.5974 & 0.6124 & 0.4628 \\
Sex M      & 23{,}559 & 0.2415 & 0.6042 & 0.5885 & 0.4550 \\
Ambulatory obs.        & 323    & 0.0899 & 0.5333 & 0.7361 & 0.4093 \\
Direct emer.           & 3{,}095  & 0.2154 & 0.5756 & 0.5975 & 0.4366 \\
Direct obs.            & 1{,}682  & 0.1791 & 0.5602 & 0.7441 & 0.4556 \\
Elective               & 1{,}661  & 0.2423 & 0.6651 & 0.5923 & 0.4742 \\
EU observation         & 3{,}289  & 0.2047 & 0.5668 & 0.7468 & 0.4667 \\
EW emer.               & 19{,}020 & 0.2237 & 0.5938 & 0.5908 & 0.4450 \\
Observation admit      & 11{,}132 & 0.2356 & 0.5838 & 0.5756 & 0.4414 \\
Surgical same-day adm. & 4{,}186  & 0.3643 & 0.6912 & 0.6228 & 0.5358 \\
Urgent                 & 4{,}644  & 0.2724 & 0.5996 & 0.5809 & 0.4686 \\
\bottomrule
\end{tabular}
\caption{R-GEAN performance by age band, sex, and admission type on the test
split (internal stress tests).}
\label{tab:subgroup}
\end{table*}

\subsection{Removal-label composition}

True removals are dominated by anti-infectives (38.2\%) and cardiovascular
agents (20.9\%). Among medication orders active at 24 hours, 26.1\% have recorded durations of
no more than one day. To test whether the removal advantage is merely due to
short courses ending, we performed a post-hoc stratification of removal
classes using their median anchor-order duration on the test split
(Table~\ref{tab:remdur}). This grouping
was not used for training, threshold selection, or model selection. R-GEAN's removal-F1 advantage over RETAIN-FullSet holds for both short- and
longer-duration classes and is in fact larger for the latter, so it is not an artifact of short orders
(GAMENet-FullSet removal F1 is 0.599 and 0.256 for the two strata).

\begin{table}[!tbp]
\centering
\small
\setlength{\tabcolsep}{5pt}
\begin{tabular}{@{}lrrr@{}}
\toprule
\textbf{Duration stratum} & \textbf{Classes} & \textbf{R-GEAN} & \textbf{RETAIN} \\
\midrule
Median order $\leq 48$h & 27 & 0.6925 & 0.5937 \\
Median order $>48$h     & 51 & 0.4769 & 0.2774 \\
\bottomrule
\end{tabular}
\caption{Removal F1 by class order-duration stratum. Classes are grouped by
their median anchor-order duration on the test split for this post-hoc
diagnostic; the grouping was not used for training, threshold selection, or
model selection.}
\label{tab:remdur}
\end{table}

\subsection{Snapshot-event coverage}

The benchmark labels capture net differences between the 24-hour and discharge
snapshots rather than every intervening medication event. Table~\ref{tab:coverage}
classifies all 708{,}149 medication starts and stops between these landmarks.
Represented additions and removals directly alter snapshot membership; the
remaining events are transient exposures, stop--restart sequences, or
opposite-direction events within classes that still receive a net edit label.

\begin{table}[!tbp]
\centering
\small
\setlength{\tabcolsep}{5pt}
\begin{tabular}{@{}lrr@{}}
\toprule
\textbf{Event category} & \textbf{Count} & \textbf{\%} \\
\midrule
Represented addition                         & 121{,}836 & 17.2 \\
Represented removal                          & 74{,}174  & 10.5 \\
Transient exposure                           & 238{,}068 & 33.6 \\
Stop--restart, membership unchanged          & 207{,}087 & 29.2 \\
Restart of a removal class                   & 26{,}601  & 3.8 \\
Within-window stop of an addition class      & 40{,}383  & 5.7 \\
\midrule
Total                                        & 708{,}149 & 100.0 \\
\bottomrule
\end{tabular}
\caption{Complete mutually exclusive taxonomy of medication start and stop
events occurring between the 24-hour anchor and discharge. Net snapshot labels
are directly represented by the first two rows; the final two rows are
opposite-direction transitions of edit-labeled classes with multiple
within-window episodes.}
\label{tab:coverage}
\end{table}

\end{document}